\documentclass[sigconf]{acmart}

\usepackage{array}
\usepackage{lscape}
\usepackage{multirow}
\usepackage{graphicx}
\usepackage{subcaption}
\usepackage{color}
\usepackage{fontawesome5}
\usepackage{xcolor}
\usepackage{xargs}
\usepackage{balance}
\usepackage{soul}
\usepackage{mdframed}

\usepackage{enumitem}

\newcommand*\rot{\rotatebox{90}}

\definecolor{yesColor}{rgb}{0.565, 0.773, 0.482}
\definecolor{noColor}{rgb}{0.796, 0.447, 0.416}
\definecolor{maybeColor}{rgb}{0.98, 0.73, 0.01}

\newcommand{\YES}{\textcolor{yesColor}{\faCheck}}

\newcommand{\NO}{\textcolor{noColor}{\faTimes}}

\AtBeginDocument{%
  \providecommand\BibTeX{{%
    \normalfont B\kern-0.5em{\scshape i\kern-0.25em b}\kern-0.8em\TeX}}}

\setcopyright{acmlicensed}
\copyrightyear{2024}
\acmYear{2024}
\acmDOI{XXXXXXX.XXXXXXX}

\acmConference[SSDBM]{}{July 10--12,
  2024}{Rennes, France}
\acmISBN{978-1-4503-XXXX-X/18/06}

\begin{document}

\title{AI Data Readiness Inspector (AIDRIN) for Quantitative Assessment of Data Readiness for AI}

\author{Kaveen Hiniduma}
\email{hiniduma.1@osu.edu}
\orcid{0009-0006-8516-0215}
\author{Suren Byna}
\email{byna.1@osu.edu}
\orcid{0000-0003-3048-3448}
\affiliation{%
  \institution{The Ohio State University}
  \city{Columbus}
  \state{Ohio}
  \country{USA}
}

\author{Jean Luca Bez}
\email{jlbez@lbl.gov}
\orcid{0000-0002-3915-1135}
\affiliation{%
    \institution{Lawrence Berkeley National Laboratory}
    \city{Berkeley}
    \state{California}
    \country{USA}
}

\author{Ravi Madduri}
\email{madduri@anl.gov}
\orcid{0000-0003-2130-2887}
\affiliation{%
    \institution{Argonne National Laboratory}
    \city{Lemont}
    \state{Illinois}
    \country{USA}
}

\renewcommand{\shortauthors}{Hiniduma, Byna, Bez, and Madduri}

\begin{abstract}
\textit{Garbage In Garbage Out} is a universally agreed quote by computer scientists from various domains, including Artificial Intelligence (AI). As data is the fuel for AI, models trained on low-quality, biased data are often ineffective. Computer scientists who use AI invest a considerable amount of time and effort in preparing the data for AI. However, there are no standard methods or frameworks for assessing the ``readiness'' of data for AI. To provide a quantifiable assessment of the readiness of data for AI processes, we define parameters of data readiness for AI and introduce AIDRIN (\underline{AI} \underline{D}ata \underline{R}eadiness \underline{IN}spector). AIDRIN is a framework covering a broad range of readiness dimensions available in the literature that aid in evaluating the readiness of data quantitatively and qualitatively. AIDRIN uses metrics in traditional data quality assessment such as completeness, outliers, and duplicates for data evaluation. Furthermore, AIDRIN uses metrics specific to assess data for AI, such as feature importance, feature correlations, class imbalance, fairness, privacy, and FAIR (Findability, Accessibility, Interoperability, and Reusability) principle compliance.  AIDRIN provides visualizations and reports to assist data scientists in further investigating the readiness of data. The AIDRIN framework enhances the efficiency of the machine learning pipeline to make informed decisions on data readiness for AI applications. 
\end{abstract}




\maketitle

\section{Introduction}



In the age of decision-making guided by data, the quality and readiness of datasets are critical to achieving success with artificial intelligence (AI) applications. As the reliance on AI continues to grow across various domains, the need for a complete tool to assess and ensure the suitability of datasets is important. 


The current state of data evaluation tools presents a lack of comprehensive solutions. Many tools concentrate on specific aspects of data quality
(\cite{schelter2018automating,lehmberg2020datafold,informatica,DQLearn}), while others emphasize some factors of data readiness  (\cite{DRReport,gupta2021data,holland2018dataset}).
These tools have improved our capabilities in assessing datasets. However, their disjoint nature creates challenges. Often, users should rely on multiple tools to evaluate different dimensions of data readiness, leading to inefficiencies and inconsistencies in the evaluation process. Additionally, the lack of standardized metrics and methodologies in these tools restricts the establishment of benchmarks for data assessment. As a result, there remains a need for a unified and comprehensive framework to address these drawbacks and to provide a quantitative evaluation of data readiness for AI applications.

To address this gap, we introduce AIDRIN (\underline{AI} \underline{D}ata \underline{R}eadiness \underline{IN}spector), a system designed to comprehensively evaluate datasets through a diverse range of metrics, giving an overall perspective on their readiness for AI applications. While existing data quality tools \cite{schelter2018automating,lehmberg2020datafold,informatica,DQLearn}, data readiness frameworks \cite{DRReport,gupta2021data,holland2018dataset}, and data readiness dimension-specific (e.g., fairness, FAIR (Findability, Accessibility, Interoperability and Reusability) compliance) evaluation frameworks \cite{bellamy2018ai,rocca-serra2023fair,fairassist} have made significant contributions, AIDRIN distinguishes itself by addressing the limitations in these tools. Many current systems focus on specific dimensions, leaving gaps in the overall assessment. AIDRIN bridges these gaps by incorporating a set of metrics covering traditional data quality parameters, such as completeness, duplicates, outlier evaluations, AI-related metrics such as fairness considerations, privacy measures, feature relevance metrics, class imbalance evaluations, and broader FAIR principle compliance metrics. This comprehensive approach sets AIDRIN apart, providing users with a unified system for assessing the readiness of datasets for AI applications.

AIDRIN simplifies dataset evaluation by offering user-friendly data and metadata upload capabilities and simplifying the process for data practitioners and researchers. Its standout feature is in its scoring mechanism, derived from an extensive list of metrics covering a broad range of data readiness considerations. By filling the gap in comprehensive dataset evaluation shown by existing systems, AIDRIN allows users across diverse domains to measure dataset suitability for AI applications easily and efficiently. 

In this paper, we describe the design and implementation of AIDRIN, a comprehensive data evaluation toolkit for measuring AI readiness of data. In the remainder of the paper, we describe the related work (\S\ref{sec:related-work}), definition of AI readiness (\S\ref{sec:ai-readiness}), the metrics and the AIDRIN framework (\S\ref{sec:aidrin}), and its evaluation (\S\ref{sec:eval}).



\vspace{-10pt}
\section{Related Work}
\label{sec:related-work}

In this section, we review existing toolkits and frameworks, specifically focusing on those designed for assessing the quality and readiness of data, with a particular focus on applications of AI. While the literature lacks research explicitly dedicated to data readiness for AI, we categorize the existing works into three main groups: toolkits designed for evaluating general data quality, toolkits explicitly targeting to assess data readiness, and toolkits designed to address specific domains or dimensions of readiness, such as FAIR compliance and bias. 

\noindent
\emph{Data Quality Toolkits.}
Within the area of data quality toolkits, Informatica's Data Quality Tool \cite{informatica} is an open-source solution designed for data profiling, cleansing, and monitoring capabilities. It provides metrics for data completeness, accuracy, consistency, and reliability. Similarly, DQLearn~\cite{DQLearn} stands out as a framework designed for organized quality operations. It systematically addresses four main tasks: detecting data quality issues with the aid of metrics, correcting identified problems, implementing rules for custom assessment using metrics, and explaining data quality aspects. In designing AIDRIN, we incorporate a comprehensive range of metrics covering not only traditional data quality parameters but also offering assessments of AI readiness using metrics such as data bias, privacy, feature relevance, correlation, and FAIR compliance.

Deequ \cite{schelter2018automating} is another data quality assessing library designed for defining ``unit tests for data,'' enabling the measurement of data quality in large datasets. It allows users to set explicit assumptions about their data, such as attribute types, absence of null values, and more, in the form of data quality checks. These checks can then be verified on a sample of data to identify errors by defining checks on various properties of data early in the data pipeline, before feeding it to consuming systems or machine learning algorithms. In contrast, AIDRIN stands out by offering a comprehensive approach to ensure the readiness of data for machine learning tasks, encompassing a wider range of considerations beyond data quality checks alone. Additionally, in AIDRIN we offer a user-centric approach in which the users can select their assessment criterion and easily visualize the assessment results for improved interpretation.

\noindent
\emph{Data Readiness for AI Toolkits.} 
The Data Nutrition Label \cite{holland2018dataset} framework offers a standardized format for presenting essential information about datasets, including metadata, provenance, variable descriptions, statistics, pair plots, synthetic data generated from probabilistic models, and ground truth correlations. 
Gupta et al. \cite{gupta2021data} 
designed a toolkit to provide automated explanations across various dimensions of data quality, such as metadata data information, data provenance, simple statistics of data features, and linear correlations between features to simplify data preparation and improve the quality of training data for AI. 
To accommodate varying technical skills among practitioners, the design prioritizes user-friendly features and visualizations. The toolkit increases overall productivity by allowing data specialists to efficiently examine and choose datasets to accelerate AI model building and deployment. 
Data Readiness Report \cite{DRReport} focuses on addressing challenges in data preprocessing within machine learning pipelines. The proposed solution generates helpful documentation for a data quality and readiness assessment. This report offers detailed insights into data quality across standardized dimensions, documenting properties, quality issues, and data operations by various individuals. 


We designed AIDRIN to quantify readiness factors and produce relevant visualizations, significantly simplifying the analysis process by reducing the time and effort required by data specialists. 
Notably, AIDRIN's focus on addressing emerging concerns like privacy, fairness, and FAIR compliance assessments emphasizes its commitment to overcoming evolving challenges within the field. This approach not only enhances the efficiency and effectiveness of data analysis but also highlights AIDRIN's goal of promoting responsible and ethical practices in data-driven challenges.

\noindent
\emph{Domain Specific Toolkits. }
In the recent commitments towards obtaining fairness in AI models, considerable attention has been directed towards the evaluation of the fairness of data before its application in AI systems. One of the tools addressing this aspect is IBM's AI Fairness 360 toolkit \cite{bellamy2018ai} (AIF360), an open-source software designed to overcome biases in ML models. 
The toolkit incorporates bias detection metrics such as representation and statistical rates of sensitive attributes, allowing users to assess and address biases related to their specific circumstances. 

Recent research has focused on providing FAIR (Findable, Accessible, Interoperable, and Reusable) data, with a focus on developing toolkits that evaluate the level of FAIR compliance in data. 
FAIR Cookbook \cite{rocca-serra2023fair} addresses challenges in implementing FAIR principles in scientific data management, recognizing the nature of FAIR principles and the absence of a formalized standard. 
The cookbook evaluates the costs and benefits associated with FAIR data, highlighting the value of operational changes in delivering FAIR data and services. In parallel, FAIRassist \cite{fairassist}, an educational component of the FAIR sharing resource, guides users on how to measure and enhance the FAIR compliance of digital objects. 
ESS-DIVE \cite{cholia2024essdive} is a data repository for earth sciences data, which evaluates the FAIR principle compliance for the data being published. 
ESS-DIVE uses quantitative assessments based on metrics such as search performance, metadata completeness, and user feedback.

\begin{table}[!tb]
    \footnotesize
    \caption{Data Quality Assessment Tools}
    \label{tab:data_quality_tools}
    \centering
    \begin{tabular}{lccccccccc}
        \textbf{Tool} & \rot{\textbf{\scriptsize Completeness}} & \rot{\textbf{\scriptsize Outliers}} & \rot{\textbf{\scriptsize Duplicates}} & \rot{\textbf{\scriptsize Privacy}} & \rot{\textbf{\scriptsize Fairness}} & \rot{\textbf{\scriptsize FAIR Compliance}} & \rot{\textbf{\scriptsize Feature Correlations}} & \rot{\textbf{\scriptsize Feature Relevancy}} & \rot{\textbf{\scriptsize Class Imbalance}} \\
        \toprule
        Informatica \cite{informatica} & \YES & & & & & & & &\\
        \midrule
        DQLearn \cite{DQLearn} & \YES & \YES & \YES & & & & & & \\
        \midrule
        Gupta et al \cite{gupta2021data} & \YES & \YES & \YES & & \YES & & \YES & \YES & \YES \\
        \midrule
        Data Readiness Report \cite{DRReport} & \YES & \YES & & & & \YES & \YES & & \\
        \midrule
        AI360 \cite{bellamy2018ai} & & & & & \YES & & & & \\
        \midrule
        FAIR Cookbook \cite{rocca-serra2023fair} & & & & & & \YES & & & \\
        \midrule
        FAIRassist \cite{fairassist} & & & & & & \YES & &  & \\
        \midrule
        ESS-DIVE FAIR \cite{cholia2024essdive} & & & & & & \YES & &  & \\
        \midrule
        \textbf{AIDRIN} & \YES & \YES & \YES & \YES & \YES & \YES & \YES & \YES & \YES \\
        \bottomrule
    \end{tabular}
    \vspace{-10px}
\end{table}

Despite the availability of various tools and frameworks that tackle different pieces of the AI readiness evaluation, no single system evaluates all the aspects using quantitative metrics and visualizations. 
Towards addressing this gap, we present AIDRIN, a comprehensive framework that addresses various dimensions of data readiness, including FAIR principles compliance. 
In Table \ref{tab:data_quality_tools}, we compare various features of existing tools with AIDRIN to highlight the scope and features offered by our framework. 

\vspace{-5px}
\section{Defining Data Readiness for AI}
\label{sec:ai-readiness}

The concept of ``Data Readiness for AI'' lacks a standard definition and is still evolving. 
To find a comprehensive definition of data readiness for AI, we have recently conducted a survey of existing metrics and partial definitions \cite{hiniduma2024data}. 
In our effort to design a comprehensive framework and metrics for AI readiness, we propose seven categories of evaluation and several metrics in each category. As a standard for defining AI readiness of data is still evolving, we anticipate a few changes to this classification. Given below are the categories of data readiness for AI.

\vspace{5pt}
\noindent
\textbf{Categories of Data Readiness for AI}

\begin{itemize}[leftmargin=*]
    \item \textbf{Quality}
\begin{itemize}
    \item \textbf{Completeness}: Measures the presence of all required data.
    \item \textbf{Outliers}: Identifies anomalous data points that deviate from the norm.
    \item \textbf{Duplication}: Evaluates the presence of duplicate records.
    \item \textbf{Data preparation practices}: Assesses the robustness of the methods used to prepare the data.
    \item \textbf{Timeliness}: Ensures data is up-to-date and relevant.
\end{itemize}

\item \textbf{Understandability}
\begin{itemize}
    \item \textbf{Metadata availability and quality}: Ensures comprehensive metadata is present to describe the dataset.
    \item \textbf{Provenance}: Tracks the origin and lineage of the data, ensuring accuracy and completeness.
    \item \textbf{User interfaces for data access}: Evaluates the ease of accessing and interacting with the data.
\end{itemize}

\item \textbf{Structural quality}
\begin{itemize}
    \item \textbf{Used data types}: Assesses the appropriateness and consistency of data types.
    \item \textbf{Quality of data schema}: Evaluates the design and structure of the data schema that supports normalized forms and fast data storage and access.
    \item \textbf{File format and used data storage system}: Reviews the efficiency and suitability of file formats and storage systems.
    \item \textbf{Data access performance}: Measures the speed and reliability of data retrieval.
\end{itemize}

\item \textbf{Value}
\begin{itemize}
    \item \textbf{Feature importance}: Assesses the significance of different features within the dataset.
    \item \textbf{Labels}: Examines the availability, quality, and correctness of labels for supervised learning.
    \item \textbf{Data point impact}: Evaluates the influence of individual data points on the overall dataset.
    \item \textbf{Uncertainty in data}: Measures uncertainty or confidence in the data using uncertainty quantification methods.
\end{itemize}

\item \textbf{Fairness and bias}
\begin{itemize}
    \item \textbf{Class imbalance}: Assesses the distribution of classes within the dataset.
    \item \textbf{Class separability}: Measures how well different classes can be distinguished.
    \item \textbf{Discrimination index}: Identifies potential biases in the data.
    \item \textbf{Population representation}: Ensures diverse and representative sampling of the population.
\end{itemize}

\item \textbf{Governance}
\begin{itemize}
    \item \textbf{Collection}: Reviews consent, sampling methods, ethical considerations, regulatory compliance, and funding sources.
    \item \textbf{Processing and curation}: Assesses anonymization, curation, and de-identification methods used.
    \item \textbf{Application}: Evaluates usage restrictions and potential biases in data analysis.
    \item \textbf{Security}: Reviews data sensitivity, access control mechanisms, and sharing protocols.
    \item \textbf{Privacy}: Assesses privacy requirements, budgets, and scores.
\end{itemize}

\item \textbf{AI application-specific metrics}
\begin{itemize}
    \item \textbf{Model-specific metrics}: Evaluates metrics specific to the AI models and their intended applications, ensuring the data meets the requirements for successful model training and deployment. Users can define data metrics that are specific to the AI model they are developing. 
\end{itemize}
\end{itemize}




\vspace{-5px}
\section{AI Data Readiness Inspector (AIDRIN)}
\label{sec:aidrin}

Based on the definition of AI readiness of data described above, we designed AIDRIN as a comprehensive framework to provide data assessment metrics. 
In this section, we provide the details of metrics and visualizations provided in AIDRIN. We highlight the specific metrics integrated to assess data readiness across various dimensions. Among the categories described in Section \ref{sec:ai-readiness}, we provide Quality, Understandability (using FAIR principle compliance), Value, Fairness, and Bias metrics in AIDRIN. While the privacy aspect of Governance is already integrated into AIDRIN, other components addressing Governance and Structural Quality are currently in development.

\vspace{-10pt}
\subsection{Analysis Capabilities of AIDRIN}

\begin{figure} 
  \centering
  \includegraphics[width=\columnwidth]{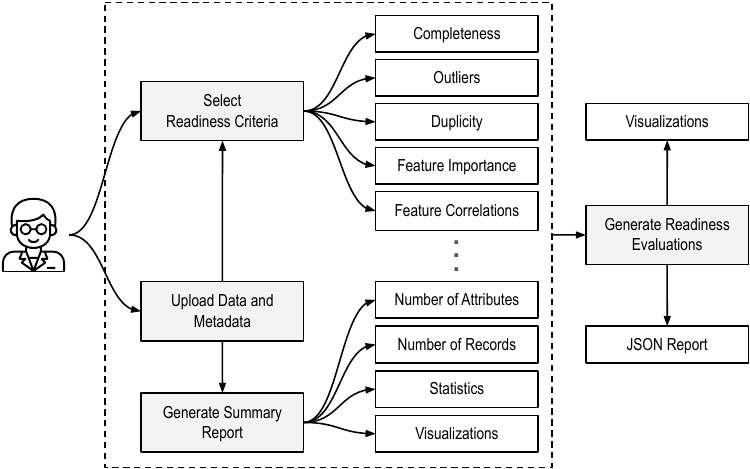}
  \caption{An overview of the AIDRIN workflow. The AI readiness assessment starts by processing data and metadata to generate summary statistics. Users can then select metrics across different data readiness dimensions for evaluation. AIDRIN generates corresponding visualizations and a comprehensive report of the selected metrics, aiding in data readiness analysis and decision-making.}
  \label{fig:system_design}
\end{figure}

In Figure \ref{fig:system_design}, we highlight the current design of AIDRIN.
For the data and metadata users enter as input, AIDRIN analyzes and provides three types of inspection results: summary statistics of data, data readiness metrics, and visualizations. The summary statistics provide general information regarding the data, such as the number of attributes and data records, statistics such as percentiles, min, max, average, standard deviation, and distribution. Users can then dive deep with selected attributes (features) to inspect a plethora of AI readiness metrics, as shown in Table \ref{tab:data_quality_tools}. We demonstrate metrics with corresponding plots and visualizations. Each metric also accompanies the definition and descriptions for ease of understanding. We describe the various metrics and visualizations below.

\subsubsection{Completeness.}
Completeness signifies the existence of necessary data and attribute values within a dataset, indicating the extent to which data points or entries fully exist with relevant attribute values. In this study, we use the completeness metric proposed by Blake et al.~\cite{blake_completeness} to assess this dimension of readiness. As defined by Blake et al., completeness refers to the presence of missing values in a dataset. AIDRIN quantifies this by measuring the proportion of missing values within each feature of the dataset.

\vspace{-10px}
\subsubsection{Outliers}
An outlier in a dataset denotes a data point or instance that deviates significantly from the anticipated values within the dataset. There are other metrics discussed in the literature to evaluate outliers like Local Outlier Factor (LOF) by Breunig et al. \cite{breunig2000lof} and Incremental Local Outlier Factor (ILOF) by Pokraja et al. \cite{pokrajac2007incremental}.
Although LOF and ILOF methods have been effective in identifying outliers by leveraging local density information, the Interquartile Range (IQR) method appears to be a better choice under various circumstances. The IQR method is a statistical tool used to assess the dispersion of data in a dataset. It centers on the middle $50\%$ of the data, calculated as the range between the first quartile and the third quartile. By computing the IQR as the difference between Q1 and Q3, the method establishes bounds for potential outliers. Data points lying beyond these bounds, determined by a user-defined constant $k$ (usually $1.5$), are identified as outliers. This approach is especially valuable when dealing with datasets open to extreme values that could distort traditional measures of central tendency. The resulting outlier score, ranging from $0$ to $1$, quantifies the proportion of outliers in each feature of the dataset, providing a robust assessment of data variability.

The IQR method is more reliable for detecting outliers because it is less affected by extreme values than the LOF method.  The simplicity and interpretability of the IQR method, based on quartiles, make it more accessible to both experts and non-experts in statistics. Additionally, the IQR method's applicability to non-normally distributed data and its consistent performance across diverse datasets highlight its flexibility. Unlike LOF, which may require careful parameter tuning, the IQR method is parameter-free, reducing sensitivity to parameter choices. These advantages position the IQR method as a reliable and straightforward option for outlier detection, particularly when dealing with datasets with varied distributions and potential outliers. Hence, AIDRIN uses IQR to assess the proportion of outliers in each feature of the dataset. 

\begin{figure}[t]
  \centering
  \includegraphics[width=0.45\textwidth]{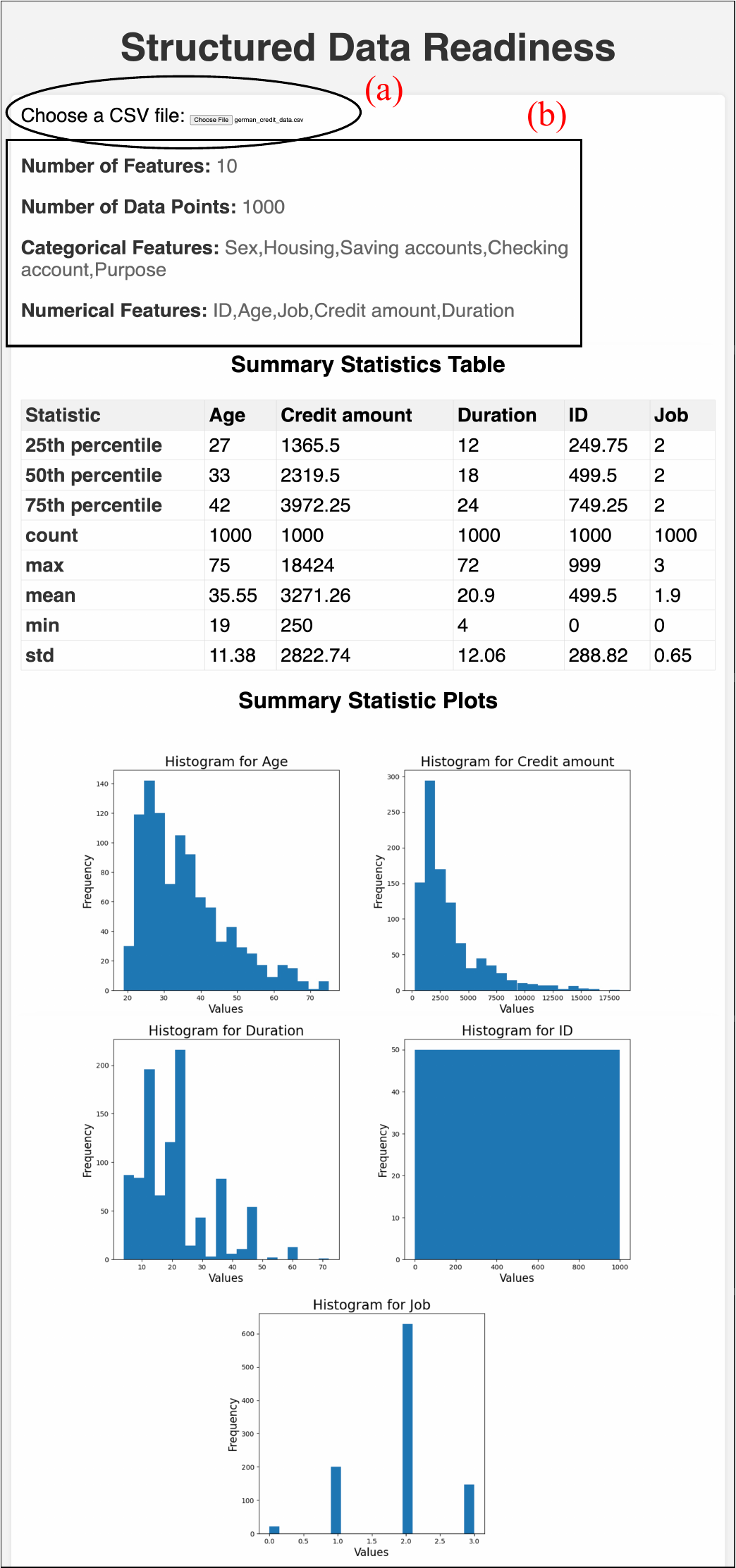} 
  \caption{Interface for uploading files and summarizing dataset statistics in AIDRIN. Users can import datasets and view key statistical summaries, the dimensions of the dataset, and the numerical and categorical features.}
  \vspace{-10px}
  \label{fig:subfig1}
\end{figure}

\begin{figure}
  \centering
  \vspace{-5px}
  \includegraphics[width=\linewidth]{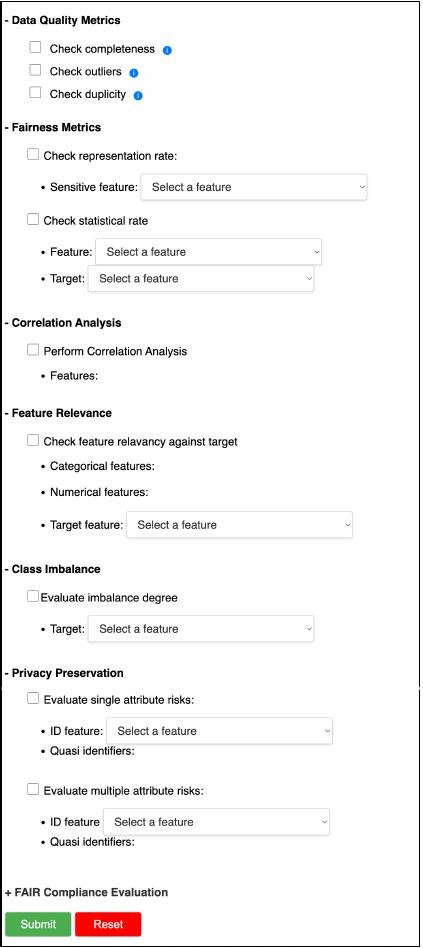} 
  \caption{
  AIDRIN enables a tailored evaluation of the data and models through customized analysis.}
  \vspace{-30pt}
  \label{fig:subfig2}
\end{figure}

\subsubsection{Duplicates}
Duplicates in a dataset refer to the existence of duplicate or redundant instances, potentially distorting the AI modeling process by introducing repetitive data entries. Bors et al.~\cite{10.1145/3190578} introduce a scoring system for detecting duplicate entries that rely on uniqueness. A score of $1$ (true) is assigned to values with a unique combination in the chosen columns, while a score of $0$ (false) is assigned to values appearing multiple times. AIDRIN uses this method to examine the number of identical items present in datasets. By comparing the number of unique items to the total number of items, it generates a single score indicating the level of duplication throughout the dataset.

\subsubsection{Privacy}
A privacy measure is a metric or method used to quantify the level of privacy protection in a given context, especially in the handling and release of sensitive data. It aims to assess the risk of re-identification of personally identifiable information in datasets. In the context of Vatsalan et al.'s work \cite{https://doi.org/10.1111/bjet.13223}, privacy is focused on reducing the re-identification risks in datasets in the education sector. The authors propose the ``MM risk score'', which uses the Markov model to quantify re-identification risks. Unlike existing approaches that often assume prior knowledge, this method considers all available information in the dataset, incorporating event-level details that link multiple records to the same individual, as well as investigating correlations among attributes. Markov model assesses the relationships among various features present in the dataset. By analyzing the transitions between the features, it identifies the correlations across the features that could lead to potential re-identifications of individual data records in the dataset. The event-level details capture the sequential occurrences between different records of the same individual. For example, three separate features analyzed collectively could disclose sensitive information about an individual compared to considering only a single attribute. AIDRIN uses this approach for assessing and managing re-identification risks, emphasizing a more detailed analysis of dataset information to improve privacy protection.

Comparing this to other privacy metrics in the literature, Vatsalan et al.'s method offers distinct advantages. Unlike metrics such as SHAP\scalebox{0.8}{R} introduced by Duddu et al.~\cite{duddu2022shapr} and privacy risk metric by Song et al.'s \cite{song2021systematic}, which is tied to assessing privacy risks in the context of a specific AI model, Vatsalan's approach takes a more collective view. Using the Markov model, it considers complex relationships within the data, going beyond model-specific considerations. This broader perspective allows for a comprehensive evaluation of re-identification risks, making it potentially applicable to various datasets and scenarios.

Furthermore, Vatsalan et al.'s approach offers a preventive strategy compared to metrics like Attack Success Rate (ASR) introduced by Carlini et al. \cite{carlini2022privacy}, which can only be measured after model training and an attack. The dynamic nature of their method allows for potential risk mitigation strategies to be implemented during the dataset release process, improving its practical usage.
\vspace{-10px}

\subsubsection{Feature Correlations}
Correlation analysis is a fundamental statistical technique used to measure the direction and strength of the correlation between two quantitative variables. It quantifies changes in one variable correspond to changes in another, with the correlation coefficient being a crucial metric to measure the degree of their association. The coefficient ranges from $[-1, 1]$, with $1$ indicating a strong positive correlation, $-1$ indicating a strong negative correlation, and $0$ suggesting no linear relationship.

Three well-known methods for measuring correlations include the Pearson correlation coefficient \cite{freedman2007statistics}, Theil's U \cite{theil1992staticprogramming}, and Cramer's V \cite{cramer1946mathematical}. The Pearson coefficient is well-suited for continuous variables, capturing linear relationships. Theil's U, designed for categorical variables, incorporates information theory concepts, while Cramer's V extends this idea to measure association strength.

While both Theil's U and Cramer's V are suitable for categorical data, Theil's U often has an advantage. Theil's U is particularly robust due to its foundation in information theory, providing a complete measure of association. It has a scoring system, ranging from $0$ (no association) to $1$ (perfect association). Theil's U is preferred over Cramer's V for its ability to handle unequal distributions and its consideration of conditional probabilities. Theil's U is also asymmetric, meaning the association between the features X and Y may differ from that between Y and X. This is because it considers the direction of information flow and the conditional probabilities between the features. Therefore, the correlation matrices it generates will observe asymmetry as Theil's U's characteristic of considering directional relationships. In instances where categorical variables exhibit different levels of entropy, Theil's U provides more reliable and informative results.

AIDRIN utilizes the Pearson correlation coefficient to gauge the strength and direction of numerical feature correlations.  On the other hand, for the assessment of categorical feature correlations, AIDRIN uses Theil's U. This dual approach reflects the flexibility of AIDRIN's correlation analysis, enabling a more meaningful examination of the complex inter-dependencies present in the data.

\subsubsection{Fairness and Bias}

\begin{figure}
  \centering
  \vspace{-5pt}
  \includegraphics[width=0.45\textwidth]{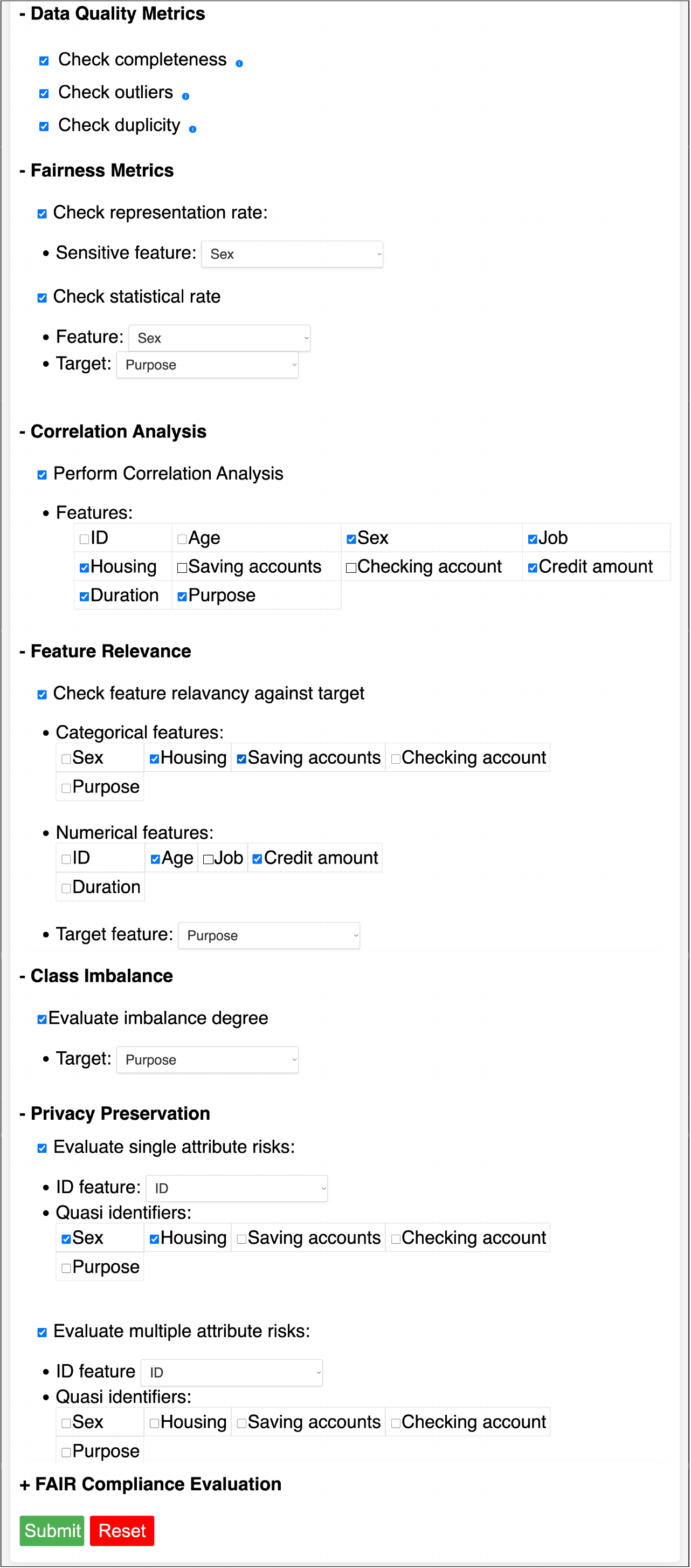} \hfill
  \caption{AIDRIN user interface for selecting metrics for the German Credit Datasets \cite{hofmann1994statlog} dataset.}
  \label{fig:ui_fig}
  \vspace{-15pt}
\end{figure}

Fairness in a dataset refers to the equal treatment of different groups or individuals, particularly concerning sensitive attributes such as race, gender, or socioeconomic status. According to Mehrabi et al. \cite{mehrabi2022survey} unfair biases in datasets can lead to discriminatory outcomes when machine learning models are trained on them, potentially extending and increasing existing societal inequalities. To ensure ethical AI applications, it is necessary to assess and address these biases in the data.

Several metrics have been proposed to quantify and evaluate fairness in datasets. The Difference and P-Difference metrics, introduced by Azzalini et al. \cite{azzalini2022eFairDB}, compare the confidence of a dependency with and without the consideration of sensitive attributes to evaluate bias, identifying important attributes contributing to unfairness.

\begin{figure*}
  \centering
  \includegraphics[width=.95\linewidth]{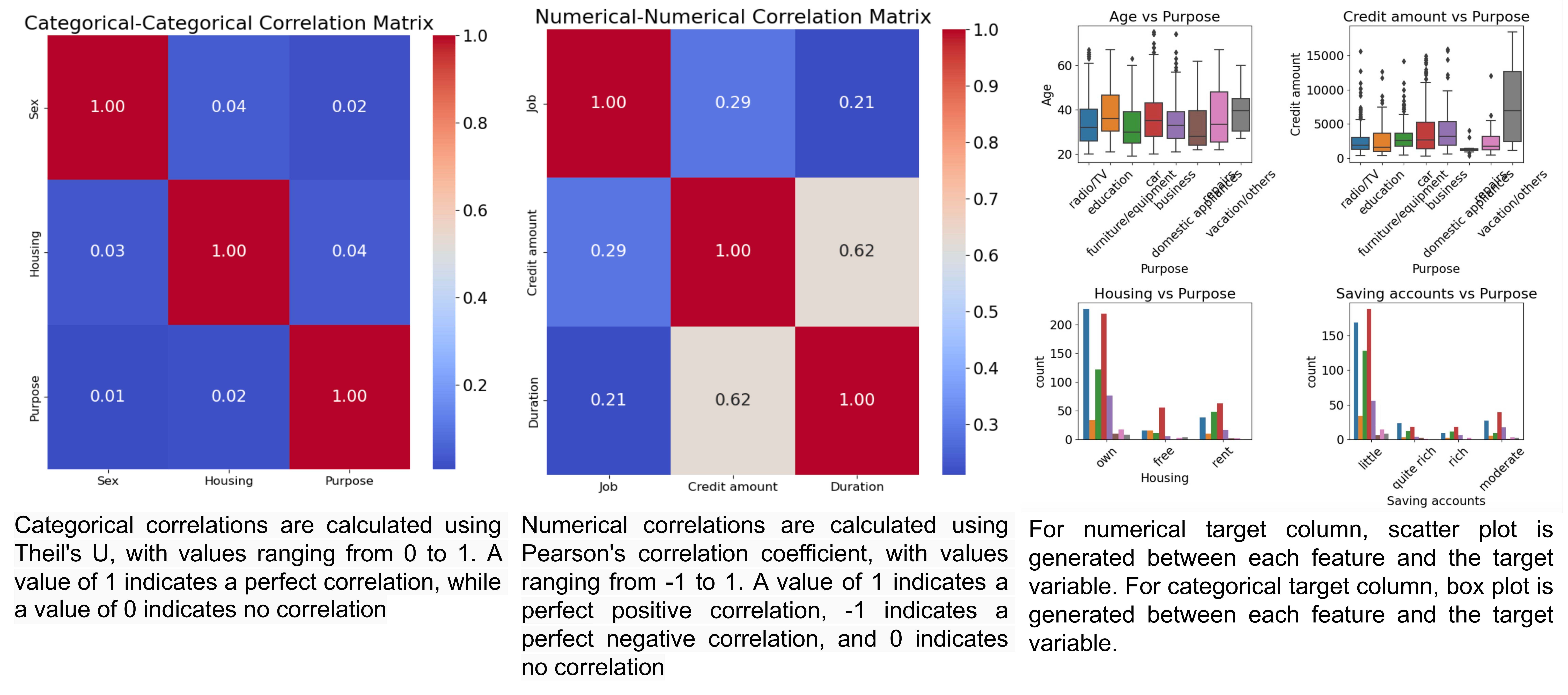}
  \vspace{-5pt}
  \caption{A sample of the visualizations generated by AIDRIN based on user-selected metrics during SGC dataset evaluation.}
  \label{fig:out_vis_fig}
\end{figure*}

Feldman et al. \cite{feldman2015certifying} introduced the metric Likelihood Ratio ($LR_+$), which assesses the disparate impact in a dataset based on specificity and sensitivity. Celis et al. \cite{celis2020datapreprocessing} contributed two metrics: the representation rate, which quantifies fairness in representing different attribute values, and the statistical rate, where the fairness evaluation is conducted by analyzing the conditional probabilities of class labels based on attribute values.

In the context of AIDRIN, the representation rate and statistical rate emerge as effective metrics for measuring group fairness. The representation rate assesses the distributions of different sensitive attributes in the dataset. The statistical rate, on the other hand, evaluates fairness through conditional probabilities, ensuring that the target representations do not discriminate certain groups based on sensitive attributes. In addition to their effectiveness in measuring group fairness, the representation rate and statistical rate metrics stand out for their ease of understanding and visualization. These qualities contribute to their practical advantage in guiding decision-makers and stakeholders toward a clearer understanding of the fairness issues within the dataset.

However, many traditional unified statistical rates are limited to binary-sensitive attributes, leaving non-binary attributes unaddressed. To tackle this limitation, we introduce the Target Standard Deviation (TSD) metric. This metric goes beyond binary groups by considering the average differences across all possible subgroups for a given target. Capturing the average differences across all sensitive groups for a given target considers variations that may exist within non-binary attributes. Moreover, its unified scalar value facilitates straightforward and interpretable evaluations, enabling stakeholders to compare and prioritize decisions effectively. However, the TSD metric may not capture contextual factors and proxies of the sensitive attribute within the dataset. It also cannot consider multiple sensitive attributes simultaneously, limiting its applicability.

Consider the categorical sensitive attribute provided by the user (that exists in the dataset) as $A \in [1,N]$ with $N$ possible values (e.g., ethnicity), and let $Y \in [Y_1,Y_2,Y_3 ...]$ represent the targets, respectively. To formulate the TSD of $Y_1$:

\begin{equation}
\text{TSD} = \sqrt{\frac{1}{N} \sum_{n=1}^{N} (Pr\left( Y = Y_1 | A = n \right) - \mu)^2}
\end{equation}
Here, $\mu$ denotes the average target probability across all groups. A lower TSD for a particular target suggests that the groups have similar treatment within that target category. Poulain et al. \cite{poulain2023improving} introduced this concept, where their metric evaluates algorithmic fairness across different sensitive groups by computing the standard deviation of the groups' true positive rates.

\subsubsection{FAIR Principle Compliance}

\begin{figure}
  \centering
  \vspace{-5pt}
  \includegraphics[width=\linewidth]{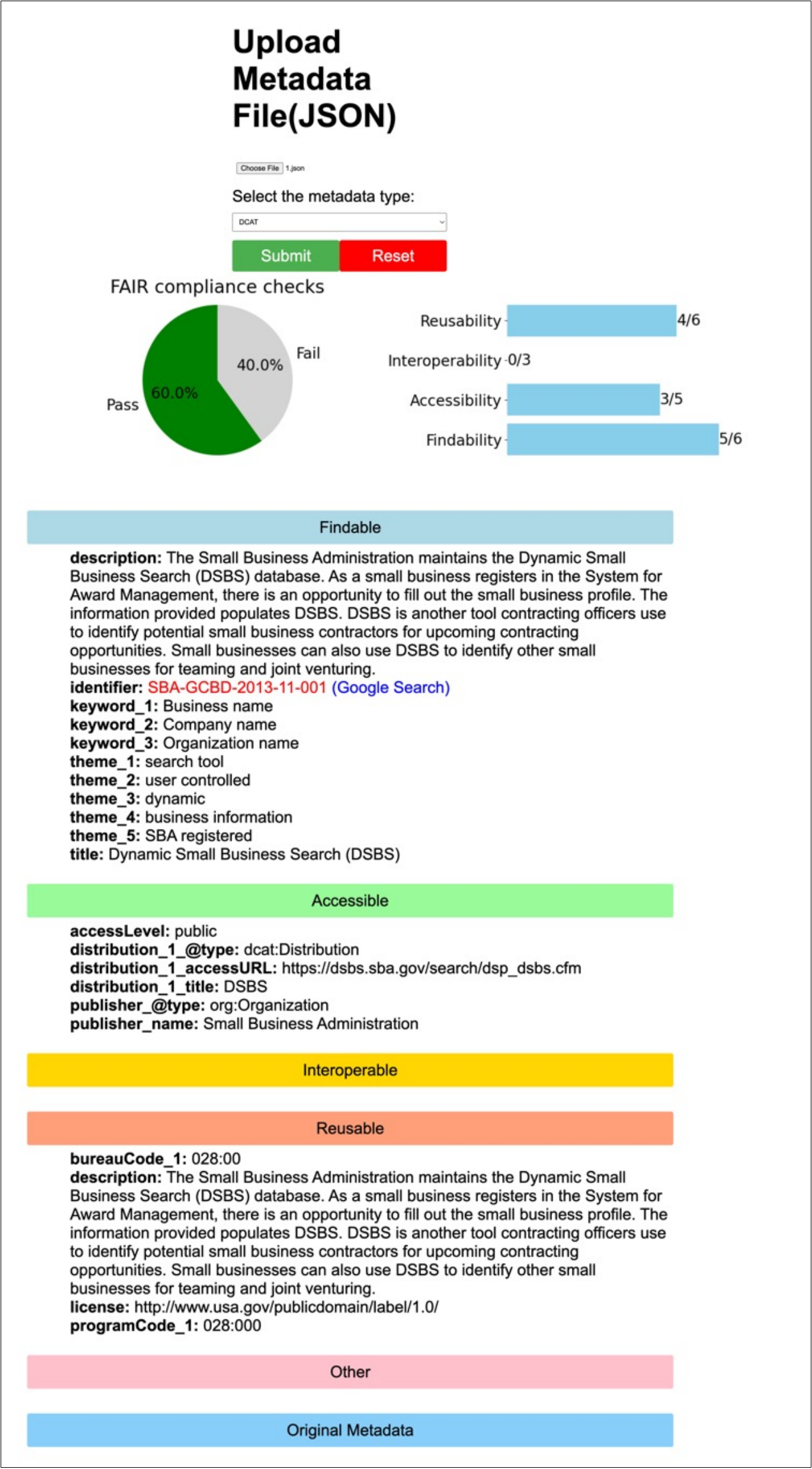}
  \caption{Illustration demonstrating the compliance check for FAIR principles applied to a DCAT data catalog.}
  \label{fig:FAIR_vis}
  \vspace{-15px}
\end{figure}

Metadata plays a crucial role in managing data. FAIR principles \cite{GOFAIR2022} have emerged to guide data publishers to improve the quality of data. 
AIDRIN assesses the FAIR (Findable, Accessible, Interoperable, Reusable) compliance of a given set of metadata. 

AIDRIN currently supports the assessment of  FAIR compliance of metadata using two metadata standards: DCAT \cite{dcat-us} and Datacite \cite{datacite-schema}. In the DCAT format, the evaluation of the ``Findable'' category involves verifying the existence of essential keys associated with identification, description, and title. Special attention is given to descriptive keywords to improve metadata discoverability. Within the ``Accessible'' domain, the function checks keys related to access levels and publisher information. ``Interoperability`` is examined through keys that enhance compatibility with established standards. Under the ``Reusable'' principle, it examines keys associated with licensing and other keys that make the data trustworthy for reuse. 

In Table \ref{tab:dcatFAIR}, we show the DCAT elements divided into each subcategory under the FAIR principles \cite{GOFAIR2022}. It categorizes metadata keys into specific aspects of FAIR, counts the checks for each category, calculates a FAIR compliance score, and generates visualizations illustrating the distribution of checks across the four aspects. For Findability, AIDRIN checks for six metadata keys, while for accessibility, it examines five metadata keys. Interoperability involves three checks and the data is reusable if the four involving metadata keys under reusability are present. 
The overall FAIR compliance score is presented as a percentage, reflecting the proportion of fulfilled checks out of the total possible checks. The higher the score, the more FAIR the metadata is considered. This concept was inspired by the work conducted by Cholia et al.\cite{cholia2024essdive} in FAIR assessment of ESS-DIVE (Environmental System Science) data.
The ``Other'' category captures keys not falling into predefined FAIR categories, providing a comprehensive metadata analysis beyond the core principles. The function calculates a FAIR compliance score by assessing the fulfillment of checks against the total possible checks, presenting the results in a pie chart as illustrated in the pie chart in Figure \ref{fig:FAIR_vis}. This approach offers an overview of metadata quality in adherence to FAIR principles within the DCAT framework. Similarly, AIDRIN supports corresponding FAIR elements in the DataCite standard. The FAIR principle assessment module in AIDRIN is extendable to other metadata standards. 

\begin{table*}[htbp]
  \footnotesize
  \caption{DCAT FAIR Compliance Categorization}
  \vspace{-5px}
  \centering
  \begin{tabular}{|c|c|>{\centering\arraybackslash}p{2cm}|>{\arraybackslash}p{11cm}|}
    \hline
    \multirow{1}{*}{\textbf{Principle}} & \textbf{Subcategory} & \textbf{Elements} & \textbf{Reasoning} \\
    \cline{2-4}\hline
    & F1 & identifier & Uniquely identifies the metadata, ensuring global uniqueness and persistence. \\
    \cline{2-4}
    \multirow{3}{*}{\textit{Findable}} & F2 & title, description, keyword, theme & Provide rich metadata, making it easier for users and computers to understand and find relevant datasets. \\
    \cline{2-4}
    & F3 & identifier & Reiterating the importance of explicitly linking metadata with the data they describe. \\
    \cline{2-4}
    & F4 & landingPage & Registering and indexing metadata in a searchable resource, enhancing discoverability. \\
    \hline
    \multirow{6}{*}{\textit{Accessible}} & A1 & distribution, downloadURL & Enable retrieval of data using a standardized protocol, providing a consistent and interoperable access mechanism. \\
    \cline{2-4}
    & A1.1 & format & Ensures openness, freedom, and universality in implementing the communication protocol. \\
    \cline{2-4}
    & A1.2 & accessLevel & Include information about authentication and authorization procedures, ensuring secure access when necessary. \\
    \cline{2-4}
    & A2 & publisher & The publisher information serves as a reference point for users seeking information about the dataset, even if the data is no longer accessible. \\
    \cline{2-4}
    \hline
    \multirow{3}{*}{\textit{Interoperable}} & I1 & format, conformsTo & Contribute to formal, accessible, and broadly applicable knowledge representation. \\
    \cline{2-4}
    & I2 & format, conformsTo & Support the use of vocabularies aligned with FAIR principles, ensuring compatibility and adherence to standards \\
    \cline{2-4}
    & I3 & references & Allows metadata to include qualified references, fostering interoperability by linking related datasets. \\
    \hline
    \multirow{8}{*}{\textit{Reusable}} & R1 & format, description, license & Contribute to rich metadata descriptions, supporting optimal reuse. \\
    \cline{2-4}
    & R1.1 & license & Ensures that data are released with clear and accessible licensing information, promoting understanding and adhering to usage terms. \\
    \cline{2-4}
    & R1.2 & programCode, bureauCode & Provides detailed information about the origin and history of the data, enhancing transparency and trust in data reuse. \\
    \cline{2-4}
    & R1.3 & conformsTo & Ensures that metadata and data follow domain-relevant community standards, facilitating integration and reuse within specific communities. \\
    \cline{2-4}
    \hline
  \end{tabular}
  \label{tab:dcatFAIR}
\end{table*}
\vspace{-5pt}

\subsubsection{Class Imbalance}

Class imbalance in a dataset occurs when the number of instances in one class significantly differs from the number of instances in another class, particularly common in binary classification problems. Hassanin et al.~\cite{hasanin2019severely} highlighted that class imbalance can lead machine learning models to discriminate their predictions against the minority class. To address this, various metrics have been proposed in the literature to measure and understand the extent of class imbalance.

One such metric is the Imbalance Ratio (IR), introduced by Francisco et al. \cite{alberto2018learning}. This metric provides a numerical representation of the disparity between the majority class instances and minority class instances in a dataset. A higher IR indicates a more imbalanced dataset, while a lower IR suggests a relatively balanced distribution. Another metric, the Imbalance Degree (ID), proposed by Ortigosa-Hernández et al. \cite{ortigosa2017measuring}, extends the measurement of class imbalance by considering specific characteristics of the class distribution. However, ID has limitations, including sensitivity to the choice of distance function and potential unreliability in extreme cases. In response to these challenges, the Likelihood Ratio Imbalance Degree (LRID), introduced by Zhu et al. \cite{zhu2018lrid}, emerges as a novel metric. It uses the likelihood ratio (LR) test, offering a fine measurement of imbalance by comparing the existing class distribution to a balanced distribution. LRID aids researchers in making informed decisions on data preprocessing to mitigate the impact of class imbalance on AI models.

AIDRIN uses the ID as the chosen metric to measure class imbalance. Several factors contribute to the selection of ID over other metrics. ID considers specific characteristics of the class distribution, providing a more accurate representation of class imbalance. This advanced representation allows for a deeper understanding of the dataset's imbalance. The ID is flexible, offering a comprehensive measure of imbalance without relying on complex statistical tests. This simplicity makes it easier to understand and apply, enhancing its practical utility in various machine-learning scenarios. The code for ID is openly available, facilitating easier implementation for researchers and practitioners. This transparency encourages widespread adoption and promotes collaboration in the community. Researchers can readily access and integrate the ID metric into their workflows, easing the process of addressing class imbalance.

\subsubsection{Feature Importance}
In training AI models on datasets with numerous features, identifying the important features that contribute to the prediction task is necessary. AIDRIN uses Shapley values \cite{NIPS2017_7062} as a method to measure feature importance, a technique widely used in various existing works (Aas et al. \cite{aas2019explaining} Frye et al. \cite{frye2019asymmetric} Merrick and Taly \cite{merrick2019explanationgame}). Shapley values provide a means to quantify the impact of each feature on a model's predictions.

AIDRIN uses a systematic method to calculate Shapley values, starting with data preprocessing and cleaning to optimize it for model training. Before the preprocessing and cleaning stage, the user uploads the data and specifies the feature set and target feature. This preprocessing step is automated but not extensively rigorous. It handles tasks like managing missing values, duplicates, and outliers, as well as performing one-hot encoding for specified categorical variables. We emphasize that ensuring clean data during the initial upload stage contributes to better results. Since AIDRIN does not conduct an extensive data cleaning process, the quality of the input data greatly impacts subsequent analyses and model performance. AIDRIN currently supports classification tasks in generating the Shapely values. After the data is transformed to meet the requirements of model training, a Random Forest Regressor is trained to perform the classification task. Our focus was not entirely on maximizing utility but rather on observing the important features most likely to contribute. Therefore, since Shapley values are computed based on the contributions of the features for the prediction task, they play an important role in assessing the influence of each feature on model predictions. 

To visualize the impact of selected features on the target variable, AIDRIN offers a range of plots based on the nature of the target variable. For numerical target variables, scatter plots are generated to showcase relationships between numerical features and the target. This visualization aids in identifying patterns and trends within the data. On the other hand, for a categorical target variable, box plots are generated to illustrate the distribution of numerical features concerning different categories of the target (Figure \ref{fig:out_vis_fig} right). These box plots provide insights into the variability of features across different target categories, facilitating a deeper understanding of the feature dynamics. The framework can also generate categorical-categorical bar charts (Figure \ref{fig:out_vis_fig} right) to provide insights into the distribution of categories and their impact on the target variable. These bar charts offer a visual representation of the counts or proportions of each category within the selected features concerning the target variable.

We highlight that AIDRIN's simplified user interaction requires only uploading the data and selecting the readiness criteria. However, the more information provided to AIDRIN, the better its results can be. For instance, evaluating feature importance via Shapely values directly correlates with the thoroughness of data cleaning and preprocessing. Currently, that stage in AIDRIN is limited to removing missing records, managing duplicates and outliers, and one-hot encoding. However, users can expect to achieve a better feature importance analysis by providing thoroughly processed data.

\subsection{Implementation of AIDRIN Framework}
We have implemented the AIDRIN framework using Python. The essential Python packages required to use AIDRIN are Flask, Pandas, Matplotlib, and Scikit-learn. Flask is the web application framework that powers the AIDRIN user interface component. Pandas handles and processes the datasets efficiently, while Matplotlib enhances AIDRIN with visualization capabilities.

AIDRIN offers multiple ways for users to analyze data. The web interface, developed using HTML and JavaScript, allows users to upload datasets and start the analysis process. Web requests from the user will then launch the Flask server and generate the analysis results and visualizations.

Additionally, AIDRIN offers a PyPI (Python Package Index) package \cite{pypi_AIDRIN} to users who are proficient in Python and comfortable in using environments like Jupyter Notebooks. Users can install the AIDRIN PyPI package via the command line and use it for data readiness assessment. This package can also be used to improve the modularity of AIDRIN by allowing contributors to expand the accessibility of AIDRIN to different domains.


AIDRIN offers flexibility for users to select specific metrics applicable to their evaluation objectives. Subsequently, the system dynamically produces visualizations and scores aligned with the selected metrics. This comprehensive report serves as a valuable insight into the implementation of AI algorithms, providing users with insightful perspectives on their data.

AIDRIN allows users to download the generated report in JSON format, enabling convenient reference and sharing. Figure \ref{fig:system_design} provides a visual representation of the essential stages within AIDRIN, including data upload (Figure \ref{fig:subfig1} (a)), dimensions of the data including the numerical and categorical attributes (Figure \ref{fig:subfig1} (b)), summary statistics generation (Figure \ref{fig:subfig1}), and metric selection (Figure \ref{fig:subfig2}). Figure \ref{fig:ui_fig} illustrates the metric selection phase, after the data upload is completed, where the attribute names in the dataset are automatically extracted for the user to define the evaluation criteria. In this report, AIDRIN uses the German Credit dataset \cite{hofmann1994statlog}.

\subsection{Towards an Aggregate AI Readiness Score}

AIDRIN currently has a set of metrics to measure the readiness of data. Our goal is to aggregate a list of these metrics to develop an ``AIDRIN Score'' that summarizes the readiness of a given dataset for AI. This rating will provide a broad view of the dataset's quality, usability, and suitability for AI applications. It will enable researchers and practitioners to make informed data preparation and usage decisions. By establishing a robust and comprehensive framework, AIDRIN aims to standardize the assessment process. 
However, because the standards of data readiness for AI are still evolving, we are in the process of defining an ``AIDRIN Score''. 

\vspace{-5px}
\section{Evaluation of AIDRIN}
\label{sec:eval}

\begin{table}[!b]
    \vspace{-10px}
    \caption{Performance Evaluation of AIDRIN across diverse datasets of varying domains and sizes.}
    \label{tab:performance_evaluation}
    \footnotesize
    \setlength{\tabcolsep}{3.7pt}
    \vspace{-10px}
    \centering
    \begin{tabular}{lccccccccccc}
        \textbf{Dataset} & \rot{\textbf{Features}} & \rot{\textbf{Records}} & \rot{\scriptsize \textbf{Completeness}} & \rot{\scriptsize \textbf{Duplicates}} & \rot{\scriptsize \textbf{Outliers}} & \rot{\scriptsize \textbf{Fairness}} & \rot{\scriptsize \textbf{Feature Relevance}} & \rot{\scriptsize \textbf{Correlation Analysis}} & \rot{\scriptsize \textbf{Class Imbalance}} & \rot{\scriptsize \textbf{Privacy}} & \textbf{Time(s)}\\
        \toprule
        RPA \cite{marcinkevics2023regensburg}& 58 & 782 & \YES & \YES & \YES & \YES & \YES & \YES & \YES & \YES & 0.8 \\
        \midrule
        SGC \cite{hofmann1994statlog} & 10 & 1K & \YES & \YES & \YES & \YES & \YES & \YES & \YES & \YES & 0.98\\
        \midrule
        MIDRC \cite{midrc} & 21 & 60K & \YES & \YES & \YES & \YES & \YES & \YES & \YES & \NO & 1.5 \\
        \midrule
        SEHM \cite{marinlopez2023single} & 10 & 416K & \YES & \YES & \YES & \scriptsize{N/A} & \YES & \YES & \scriptsize{N/A} & \scriptsize{N/A} & 4.8 \\
        \midrule
        MetroPT-3 \cite{davari2023metrop3}& 17 & 1.5M & \YES & \YES & \YES & \scriptsize{N/A} & \YES & \YES & \scriptsize{N/A} & \scriptsize{N/A} & 5.3 \\
        \bottomrule
    \end{tabular}
\end{table}

We evaluate AIDRIN to test its performance, usability, and features.
To evaluate performance, which is needed for all data analysis tools, we present completion time in generating all metrics using diverse structured datasets. We have examined the functionality and performance of AIDRIN in varying scenarios, examining its outputs and visualizations. The resulting analysis provides a detailed account of AIDRIN's efficacy, interpreting its flexibility in handling five datasets (shown in Table \ref{tab:performance_evaluation} and showcasing the results that contribute to a better understanding of its capabilities. 

We have conducted a user study with the help of various groups of researchers to evaluate the usability of the tool. 
We evaluate the features of AIDRIN using two case studies, i,e., German Credit (SGC) dataset \cite{hofmann1994statlog} and Cancer Genome Atlas Lung Adenocarcinoma (TCGA-LUAD) clinical data of patients accessed from the National Cancer Institute NCI Data Portal \cite{grossman2016toward}. For SGC, we used the web interface to produce visualizations, and for the Cancer dataset, we used the Jupyter Notebook interface. We conducted all our experiments on a laptop with a $12$-core Apple M2 Max chip with $32$GB RAM. 

\vspace{-5px}
\subsection{Performance Evaluation}
To evaluate the responsiveness of data analysis tasks in AIDRIN, we used execution time as a metric. The execution time represents both UI-based and Python Notebook interfaces. 
In Table \ref{tab:performance_evaluation}, we show the datasets used in our study and an overview of AIDRIN's performance in assessing a range of data evaluation metrics.
This assessment involves four distinct datasets acquired from the UCI Machine Learning Repository \cite{kellyUCI}: the Single Elder Home Monitoring(SEHM) dataset \cite{marinlopez2023single}, the MetroPT-3 Dataset \cite{davari2023metrop3}, the Regensburg Pediatric Appendicitis (RPA) dataset \cite{marcinkevics2023regensburg}, and Statlog - German Credit (SGC) dataset \cite{hofmann1994statlog}. We also evaluated the MIDRC medical cases dataset \cite{midrc} provided by NIH. Analysis times of these datasets exhibit a diverse range, with collections of below $1000$ records to datasets exceeding $1.5$ million records, with feature dimensions ranging from $10$ to over $50$. This selection of varying-sized datasets comprehensively evaluates AIDRIN's performance, assessing its efficacy across small to large-scale usage scenarios and domains, including healthcare, transportation, and finance.

In our performance evaluation, the most time-intensive aspect of AIDRIN was identified as the computation of Shapley values to evaluate crucial features within a dataset. This is attributed to the necessity of training a random forest regressor for $100$ estimators based on user-specified features and target attributes. For instance, when compared to statistical feature relevance methods, the computation time for calculating the Shapley values alone for a single numerical feature (`age') and one categorical feature (`sex') on the SGC dataset took $1.62$ seconds, which is almost $2\times$ more. This value increases significantly as the number of features is increased. For two categorical features (`sex' and `housing') and two numerical features (`age' and `credit amount'), the Shapley value computation took $5\times$ more time compared to the statistical feature relevance computations. Conversely, from this analysis, the statistical measurements of feature relevance exhibited more efficient processing in comparison.  Moreover, the risk scores for privacy assessment exhibited increased time demands with growing dataset sizes. 

As shown in the performance evaluations detailed in Table \ref{tab:performance_evaluation}, it becomes evident that the time required to generate results gradually increases from $0.8$s to $5$s as the dataset size increases from $782$ records to $1.5$ million records. Notably, for the evaluation of privacy alone, the MIDRC datasets required around $106.36$ seconds for result generation on one attribute (`sex'). For two attributes (`sex' and `race'), the privacy evaluation computations required $195.53$ seconds to generate the results. As the data dimensions increase, the Markov model requires processing a greater number of possible states and combinations, resulting in longer processing times. 


Overall, AIDRIN performs well, even on a laptop. AIDRIN's analysis time depends on the number of tasks, number of data records, and computation resources. Improving the performance of data loading and using advanced computing resources is ongoing.

\vspace{-8px}
\subsection{Usability of AIDRIN: A User Study}

We have conducted a user study involving a diverse group of participants to enhance the presentation of evaluations in AIDRIN. This included three experts, three PhD students, one postdoctoral scholar, and three computer scientists, all with experience in developing data management tools and AI algorithms. This diverse participant group provided valuable insights that allowed us to refine AIDRIN and expand its potential applications.

Our evaluation included an initial phase focused on gathering feedback from experts and PhD students who tested the web interface of AIDRIN. Their assessments primarily centered on functionality, usability, and areas for improvement. Simultaneously, the computer scientists applied AIDRIN in their ongoing projects focused on privacy-preserving federated learning \cite{Ryu_APPFL_Advanced_Privacy-Preserving}, providing practical insights into its effectiveness for assessing data readiness. This phase was important in identifying usability challenges and including enhancements.
Based on the feedback, we improved the function of AIDRIN. These included enhancing the user interface and developing a PyPI package to integrate with Python environments, particularly Jupyter Notebooks. These enhancements were designed to address specific user needs identified during the evaluation, thereby enhancing the tool's usability and functionality.

In subsequent evaluations, AIDRIN's improved version was tested again, with a focus on validating these enhancements. The PyPI package, specifically designed based on user feedback, was evaluated during the NCI CRDC Artificial Intelligence Data-Readiness (AIDR) Challenge in 2024 \cite{CRDCInsights}, where its integration with Jupyter Notebooks facilitated effective data assessment. This phase confirmed the robustness and applicability of AIDRIN.

Key insights from the user study highlighted interest among participants in integrating AIDRIN into frameworks such as APPFL \cite{li2023appflx} for federated learning. This reflects AIDRIN's potential to enhance data readiness assessments within federated learning environments. We have successfully integrated data readiness assessment into the APPFL framework, which is available publicly~\cite{Ryu_APPFL_Advanced_Privacy-Preserving}. 

\vspace{-10pt}
\subsection{Data Readiness Evaluation}

\subsubsection{SGC Case Study dataset analysis using the web interface.}
In this case study, we present the findings obtained from the readiness evaluation conducted on the SGC dataset. AIDRIN summarized the data initially by displaying the dimensions of the data along with a detailed description of the summary statistics (Figure \ref{fig:subfig1}) of the existing features in the data. 

\begin{figure}
  \centering
  \vspace{-5pt}
  \includegraphics[width=.65\linewidth]{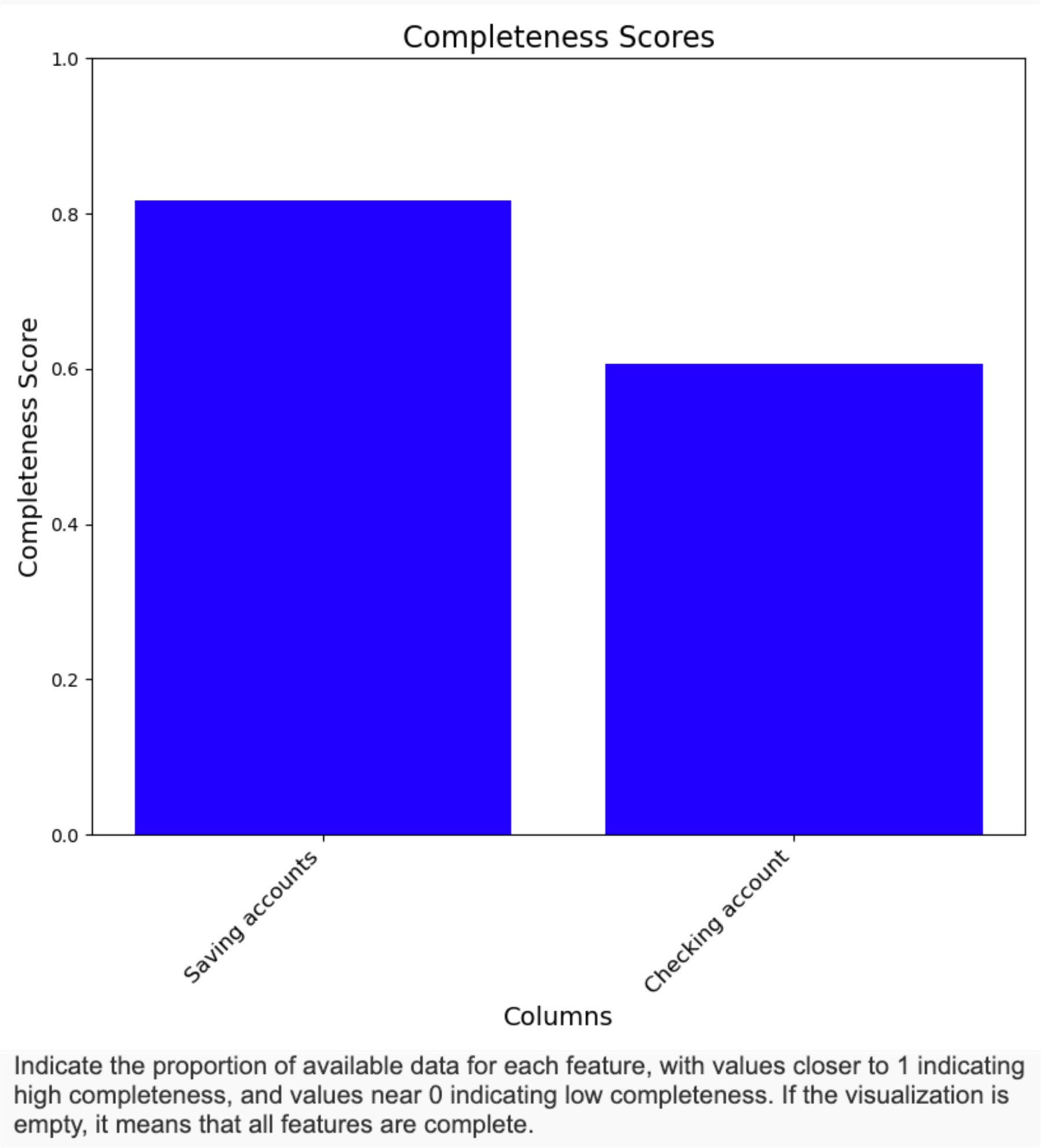}
  \caption{Visualization generated by AIDRIN of the completeness metric for the SGC dataset, illustrating the percentage of completeness for each feature.}
  \label{fig:completeness}
  \vspace{-15px}
\end{figure}

\begin{figure}
  \centering
  \vspace{-5pt}
  \includegraphics[width=.75\linewidth]{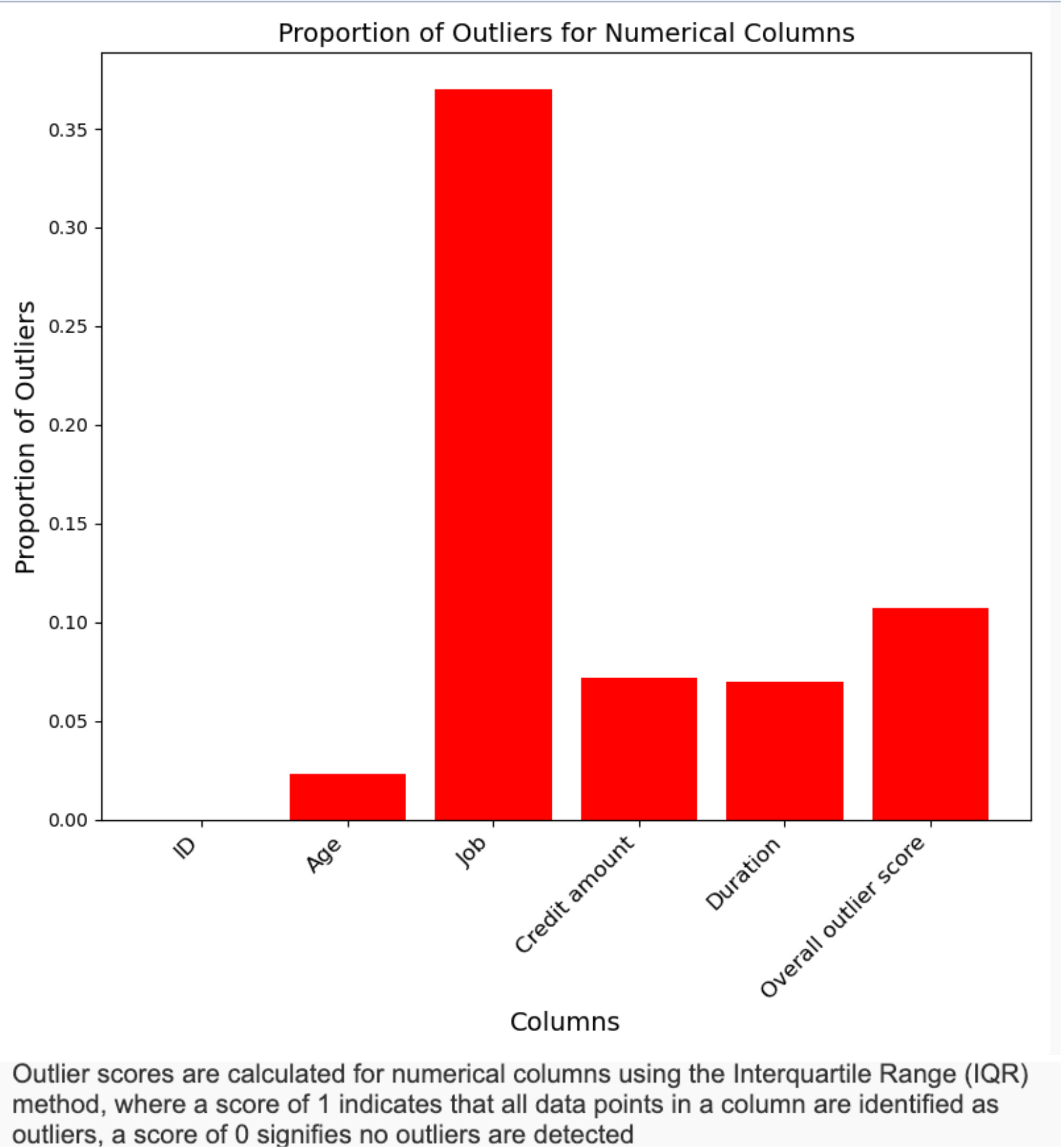}
  \caption{Visualization of outliers metric for the SGC dataset, showing the percentage of outliers for each numerical feature and the overall outlier score.}
  \label{fig:outliers}
  \vspace{-5px}
\end{figure}

After selecting the readiness metrics for the SGC dataset according to Table \ref{tab:performance_evaluation} and selection in Figure \ref{fig:ui_fig}, in terms of completeness (Figure \ref{fig:completeness}), we observed that two features were around $80\%$ and $60\%$ complete respectively. In contrast, the remaining features were fully complete. The number of duplicates in the dataset was 0, meaning there were no duplicate records in the data. Regarding outliers (Figure \ref{fig:outliers}), apart from one feature ($35\%$), the numerical features exhibited less than $15\%$ outliers. 

\begin{figure}
  \centering
  \vspace{-5pt}
  \includegraphics[width=.50\linewidth]{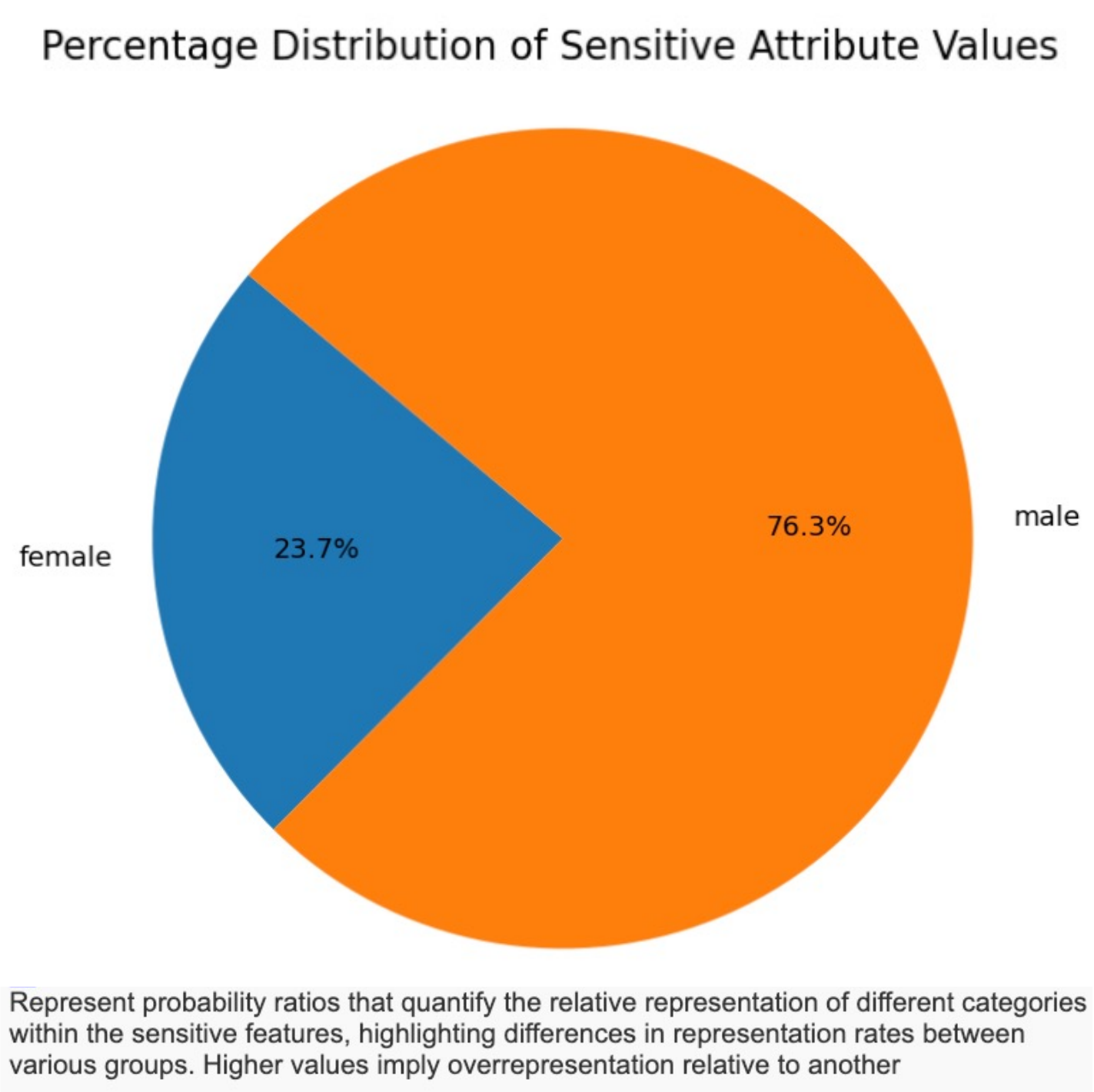}
  \caption{Visualization of representation rate metric for the SGC dataset, showing the percentage distribution of values across the selected sensitive attribute}
  \label{fig:rep_rate}
  \vspace{-15px}
\end{figure}

Fairness analysis revealed biased representation rates (Figure \ref{fig:rep_rate}), with over $75\%$ of the representation being males and the remainder being females. The class imbalance was evident when selecting the ``Purpose'' attribute, with the most popular class comprising $33.7\%$ and the minority class only $1.2\%$. The imbalance degree score of $4.49$ indicates significant class distribution disparity. When considering the privacy risk scores associated with the `Housing' feature, AIDRIN calculated a mean value of the re-identification risk scores to be $0.45$. This finding suggests the presence of a potential re-identification risk associated with the `Housing' feature.

Additionally, the visual outputs generated by AIDRIN helped to easily interpret these evaluations including other readiness evaluations such as feature relevance, and correlation analysis. Figure \ref{fig:out_vis_fig} illustrates a sample of the specific visual outputs generated based on the selections made in Figure \ref{fig:ui_fig}, enhancing the user's ability to interpret and utilize the evaluation results effectively. Additionally, users have the option to download a JSON file containing detailed evaluation results. These findings display the important role of AIDRIN in assessing the readiness of the dataset for AI applications.

\subsubsection{TCGA-LUAD Case Study dataset analysis using Jupyter Notebook.}
Another case study was conducted using the PyPI package of AIDRIN within a Jupyter Notebook. The dataset used for this case study was the Cancer Genome Atlas Lung Adenocarcinoma (TCGA-LUAD) clinical data of patients accessed from the National Cancer Institute NCI Data Portal \cite{grossman2016toward}. This readiness analysis was conducted as a part of NCI CRDC Artificial Intelligence Data-Readiness (AIDR) Challenge in 2024 \cite{CRDCInsights}. 

We observed that the overall completeness of the data was $56\%$ with attributes having varying degrees of completeness scores, the lowest being $22\%$. The number of duplicates in the data was $0\%$, ensuring that each patient's information is uniquely represented in the dataset. AIDRIN's evaluation of representation rates revealed female counterparts accounting for $65\%$ of the data records. In terms of race, $87\%$ of the individuals were identified as whites, being the majority group. This imbalance raises concerns as healthcare decisions can exhibit a disparity among different racial groups. 

Furthermore, AIDRIN observed the dataset contained a class imbalance with an ID score of $0.27$. We considered vital status as the class attribute categorized into two classes: `dead' and `alive'. This score suggests that there is a skew from a balanced class distribution that could potentially result in a skewed decision-making process. The TSD score across the gender attribute for both classes was minimal ($0.01$). Notably, the TSD score across racial groups for both classes was $0.15$. The scores reflect potential discrimination in class attributes between racial groups compared to gender groups.

The two case studies above illustrate AIDRIN's capabilities, with one using the web interface and the other using the Jupyter Notebooks. These evaluations empower scientists by providing them with valuable insights about their data to make informed decisions before proceeding to the next steps of ML pipelines. 

\vspace{-5px}
\section{Conclusion}

The importance of high-quality and AI-ready data emphasizes the success of automated decision-making. However, the data assessment process of existing tools is inefficient and inconsistent. To address these challenges, we provide a definition of data readiness parameters focused for AI and introduce AIDRIN, a comprehensive toolset designed to assess the AI readiness of data. By integrating a diverse range of metrics, AIDRIN provides users with a complete framework to assess the readiness of their data both quantitatively and qualitatively.
Moreover, AIDRIN's relevance extends into emerging areas like Federated Learning. It is important to evaluate the readiness of the edge client's data in such settings. 
We have integrated AIDRIN into the Advanced Privacy-Preserving Federated Learning (\cite{ryu2022appfl, li2023appflx,Ryu_APPFL_Advanced_Privacy-Preserving}) framework to evaluate data readiness. 

AIDRIN's current functionality is focused on tabular data, presenting a limitation in its applicability to other data modalities. Another limitation is its performance when handling huge datasets, which may impact efficiency and scalability. 
Our future work will 
cover a broader range of data modalities (e.g., images, text, and audio) as modern AI applications rely on diverse data formats. We are also designing a comprehensive scoring method that is meaningful to users in improving the readiness of data for AI.

\subsection*{Acknowledgements}

This research is supported by the U.S. Department of Energy, Office of Science, under contract numbers DE-AC02-06CH11357 (ANL), GR134836 (OSU), and DE-AC02-05CH11231 (LBNL).

\balance

\bibliographystyle{ACM-Reference-Format}
\bibliography{main}










\end{document}